\pdfoutput=1
\PassOptionsToPackage{dvipsnames, svgnames, x11names}{xcolor}
\documentclass[11pt]{article}

\usepackage[preprint]{acl}
\usepackage{booktabs}
\usepackage{bbding}
\usepackage{comment}
\usepackage[dvipsnames, svgnames, x11names]{xcolor}
\usepackage{times}
\usepackage{latexsym}
\usepackage{multicol}
\usepackage{multirow}

\usepackage[T1]{fontenc}

\usepackage[utf8]{inputenc}

\usepackage{microtype}
\usepackage{amsmath}
\usepackage{makecell}
\usepackage{inconsolata}

\usepackage{graphicx}
\usepackage[hang,flushmargin]{footmisc}

\title{BoostStep: Boosting Mathematical Capability of Large Language Models via Improved Single-step Reasoning}

\author{
Beichen Zhang$^{*1,2}$, Yuhong Liu$^{*1,2}$, Xiaoyi Dong$^{\dagger1,3}$, Yuhang Zang$^{1}$, \\ \textbf{Pan Zhang$^{1}$, 
Haodong Duan$^{1}$, Yuhang Cao$^{1}$, Dahua Lin$^{1,3}$, Jiaqi Wang$^{\dagger1 }$}\\
$^1$Shanghai AI Laboratory \quad
$^2$Shanghai Jiao Tong University \\
$^3$The Chinese University of Hong Kong \\
\texttt{\url{https://github.com/beichenzbc/BoostStep}}
}

\begin{document}

\maketitle
\footnotetext{* indicates equal contribution \par
$\dagger$ \ indicates corresponding author}

\begin{abstract}

Large language models (LLMs) have demonstrated impressive ability in solving complex mathematical problems with multi-step reasoning and can be further enhanced with well-designed in-context learning (ICL) examples. However, this potential is often constrained by two major challenges in ICL: granularity mismatch and irrelevant information.
We observe that while LLMs excel at decomposing mathematical problems, they often struggle with reasoning errors in fine-grained steps. Moreover, ICL examples retrieved at the question level may omit critical steps or even mislead the model with irrelevant details.
To address this issue, we propose BoostStep, a method that enhances reasoning accuracy through step-aligned ICL, a novel mechanism that carefully aligns retrieved reference steps with the corresponding reasoning steps. Additionally, BoostStep incorporates an effective "first-try" strategy to deliver exemplars highly relevant to the current state of reasoning. 
BoostStep is a flexible and powerful method that integrates seamlessly with chain-of-thought (CoT) and tree search algorithms, refining both candidate selection and decision-making. Empirical results show that BoostStep improves GPT-4o’s CoT performance by 4.6\% across mathematical benchmarks, significantly surpassing traditional few-shot learning's 1.2\%. Moreover, it can achieve an additional 7.5\% gain combined with tree search. Surprisingly, it enhances state-of-the-art LLMs to solve challenging math problems using simpler examples. It improves DeepSeek-R1-671B’s performance on AIME by 2.2\%, leveraging simple examples only from the MATH dataset.
\end{abstract}

\section{Introduction}
\label{sec:intro}

\begin{figure}[t]
\centering
\includegraphics[width=0.49\textwidth]{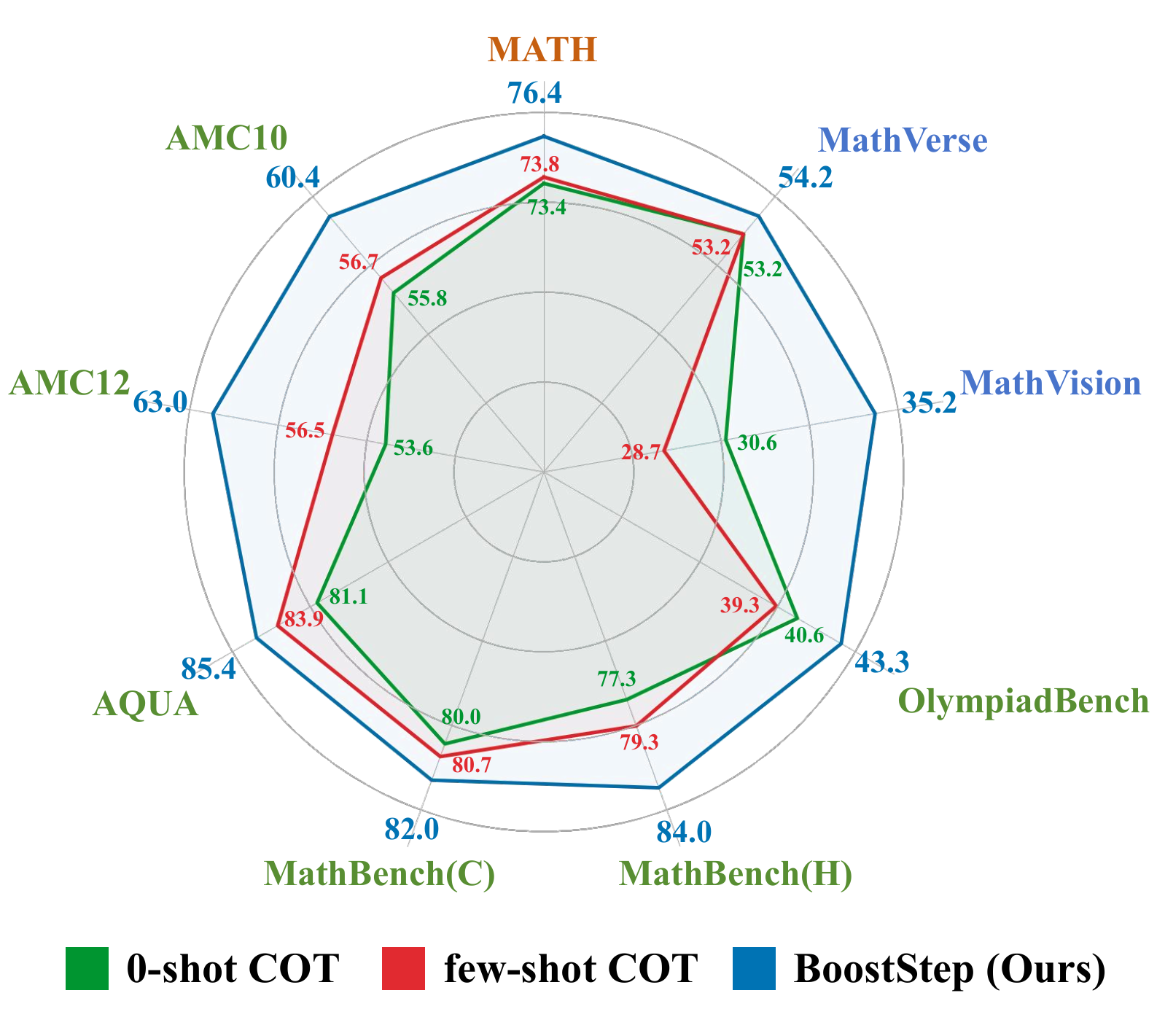}
\caption{Our step-aligned in-context learning (ICL) outperforms traditional problem-level few-shot learning for about 4\% across \textcolor{Bittersweet}{in-domain}, \textcolor{OliveGreen}{out-domain} and \textcolor{RoyalBlue}{cross-modality} mathematical benchmark on GPT4o. Moreover, on benchmarks with lower similarity with the reference problem set (i.e. OlympiadBench and multi-modal benchmarks), where problem-level ICL may have a negative impact, BoostStep still provides valuable guidance.}

\label{fig:radar}
\end{figure}

Mathematical reasoning is a crucial and challenging task in the development of artificial intelligence. It serves as an indicator of a model's ability to perform complex reasoning and has a wide range of applications, such as problem-solving, theorem proving, and scientific discovery. 

When solving complex mathematical problems, cutting-edge LLMs often adopt a multi-step reasoning strategy. Specifically, they first decompose a complex problem into several simpler steps and then solve each single step independently. 

Through the analysis of error cases, we found that current SOTA models are relatively correct in the step-dividing phase, that is, the model can know exactly what tasks should be completed in each step. However, there are still a lot of mistakes within each reasoning step, such as wrong formula use, wrong calculation, insufficient enumeration, etc. To quantitatively substantiate this observation, we provide GPT-4o-mini with a ground truth reasoning process to determine whether the error in another response was due to an overarching flawed reasoning approach or a deviation within a particular step. In less advanced models like LLaMA-3.1-8B~\cite{dubey2024llama}, 91.3\% of errors originate from single-step reasoning. In more advanced models like GPT-4o, up to 99.2\% of errors are ascribable to some particular steps. This exaggerated proportion suggests that the correctness of single-step reasoning is the bottleneck of reasoning capability.

Various approaches have been employed to improve reasoning correctness, such as producing chains of thought through prompt engineering~\cite{kojima2022large, wei2022chain}, fine-tuning with mathematical data~\cite{shao2024deepseekmath, yang2024qwen2,ying2024internlm}, or generating multiple candidate reasoning paths using Tree Search Methods~\cite{zhang2024llama, zhang2024accessing, wang2024math}. 

Among those techniques, in-context learning is a particularly important one, which offers similar examples to provide detailed guidance. However, the examples retrieved by traditional problem-level in-context learning are listed before the reasoning process, thereby lacking fine-grained guidance during the reasoning process. Moreover, since the example problem can't be identical to the new one, the irrelevant steps in those examples may even become a distraction from the current reasoning, thus even negatively affecting the single-step reasoning capability for some specific steps.

To this end, we refine in-context learning from problem-level to step-level granularity to offer similar example steps during an ongoing reasoning process for fine-grained step-aligned guidance. We also ensure that the introduced example is still relevant at the step level to avoid distractions.

Firstly, we have constructed an example problem bank with step-level granularity based on reasoning content instead of commonly adopted grammatical separation. This ensures the steps in the problem bank are consistent with the actual reasoning steps, thereby providing more appropriate guidance.

Building on the step-level granularity within the example problem bank, we propose an approach that incorporates in-context learning through a "first-try" format during an ongoing reasoning process. Specifically, for a given problem to be solved, we break down the solving process into step-by-step reasoning paths. During the reasoning of a single step, we first allow the model to attempt a `first try' to comprehend what the model currently needs to reason about. Based on this initial attempt, we searched the problem bank to find similar steps that can guide the model to accurately output the current step. This helps ensure a higher similarity between the retrieved examples and the current step so the distraction from irrelevant steps can be avoided and the guidance effect can be improved.

Compared with traditional problem-level ICL, our method provides examples during the reasoning process directly based on the steps to be solved, thereby offering more relevant guidance. It demonstrates significant improvements over traditional few-shot learning across various benchmarks, with an average increase of \textbf{3.4\%} on GPT-4o. 

Moreover, our method also reduces the sensitivity to the similarity between the example and the target problem, as two different problems can still share similar steps. Consequently, dissimilar problems can still offer effective guidance. On multi-modal benchmarks with lower similarity to example problems, traditional few-host learning has a detrimental effect, resulting in an accuracy reduction of 0.9\% on GPT-4o. In contrast, our approach still achieves an improvement of 2.8\%.

Besides, BoostStep also shows a promising potential to improve the reasoning quality on harder problems with simpler examples. With examples from MATH~\cite{hendrycks2021measuring}, it helps Deepseek-R1 achieve an improvement of 2.2\% on the much more challenging AIME problems.

Moreover, our method is also highly compatible with various current reasoning strategies that employ step-level tree search. 
Typically, a tree-search method requires a reason model to generate multiple step-level candidate reasoning paths and a critic model to evaluate the correctness of these candidates. Our approach can be integrated into both aspects. Specifically, when the reason model generates new candidate reasoning nodes, our method can introduce similar examples in the aforementioned `first-try' manner to improve the accuracy of candidates. Additionally, it can aid the critic model by incorporating similar example steps into the evaluation of candidate reasoning processes to provide similar guidance. Experiments indicate that both applications contribute positively and bring about an improvement of 8.5\% jointly on GPT-4o.

\section{Related Works}

\noindent\textbf{Mathematical Reasoning. } Mathematical reasoning has long been a highly challenging task in the field of artificial intelligence. In the early days of artificial intelligence, constrained by a lack of general capabilities, early methods~\cite{feigenbaum1963computers, fletcher1985understanding} primarily attempted to perform simple mathematical reasoning through rule-based methods. With the advent of large language models with enhanced reasoning capabilities, contemporary approaches typically focus on enhancing performance during both the training and inference phases. The first category improves mathematical capability by fine-tuning with more high-quality mathematical data~\cite{shao2024deepseekmath,yang2024qwen2,lewkowycz2022solving,yue2023mammoth,xu2024chatglm}. This strategy can fundamentally improve the base model's mathematical capabilities. However, it demands substantial high-quality mathematical data and computational resources. Consequently, more efforts have been put into exploring various techniques during inference to enhance mathematical reasoning performance. Some work~\cite{wei2022chain,kojima2022large} involves prompt engineering to enable models to generate comprehensive chains of thought. Other studies~\cite{madaan2024self,gou2023critic,ke2024critiquellm} use self-refinement techniques to revise the initial reasoning outputs.

\noindent\textbf{Step-level Mathematical Reasoning. } Recently, to further enhance mathematical reasoning capabilities, many studies have shifted the granularity of mathematical reasoning from the problem level to the step level. This approach involves addressing each next step individually and completing small segments of reasoning within the overall task. These works often employ tree searching strategies like Tree of Thoughts (ToT)~\cite{yao2024tree, besta2024graph} or Monte Carlo Tree Search~\cite{zhang2024llama,zhang2024accessing,chen2024step,feng2023alphazero, zhu2022solving}, extending multiple steps to optimize step answers and ultimately obtain the optimal solution. Additionally, Process Supervision Models (PRMs)~\cite{lightman2023let, luo2024improve} are frequently used to verify the correctness of new candidate nodes in real-time and prune reasoning paths, thereby improving the accuracy of the final answer. This more detailed auxiliary strategy demonstrates greater potential.

\noindent\textbf{In-context Learning in Mathematical Reasoning. }\ In-context learning can provide low-cost guidance to models through similar examples, thereby enhancing the quality of model outputs and their ability to follow prompts. Consequently, it has been widely adopted. However, research on in-context learning within mathematical reasoning tasks remains insufficient. Typically, this approach involves providing the model with similar problems and their ground truth solutions to offer a general strategy for solving new problems~\cite{hendrycks2021measuring, wei2022chain}. Some efforts have been made to improve the relevance of retrieved examples by designing better retrieval mechanisms and incorporating appropriate reference rejection techniques~\cite{liu2024what}. Others try to provide high-level context instead to improve the generalizability~\cite{wu2024beyond}. However, all these methods share a common limitation: the lack of fine-grained step-level guidance. Some recent approaches~\cite{dong2024progressive} introduce ICL into the reason process. However, they still perform ICL in problem granularity and thus may not offer effective guidance for next-step reasoning.

\section{Step-Level In-Context Learning}

\begin{figure*}[t]
\centering
\includegraphics[width=0.98\textwidth]{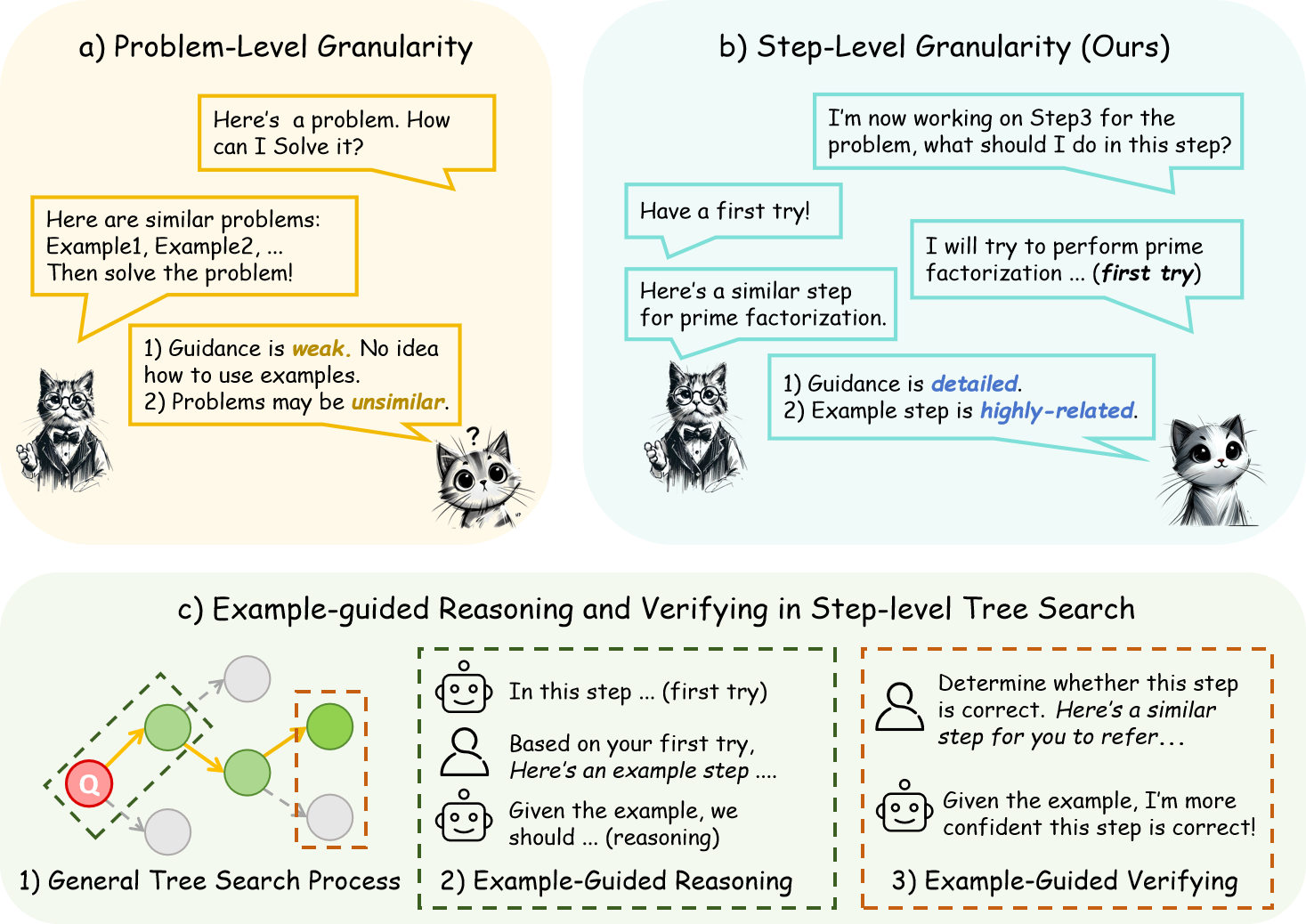}
\caption{Our strategy refines in-context learning from problem-level granularity (fig.a) to step-level granularity(fig.b) to provide more real-time fine-grained guidance. Moreover, our strategy can guide the reasoning and verifying process in tree-searching strategies by introducing examples.}

\label{fig:pipeline}
\end{figure*}
\subsection{Revisiting In-Context Learning from Conditional Probability}

Current models often employ next-token prediction for training and inference, where the conditional probability is central to the model's generation of the next token. Given a problem \( q \), a model's reasoning process can be represented by \( r_{predict} = \arg\max\limits_r P_{model}(r \mid q) \), where we train the model to get a better conditional probability $P_{model}$ so that \( r_{predict} \) can be closer to the ground truth answer \( r_{gt} 
 =\arg\max\limits_r P_{gt}(r \mid q)\). 
 
In-context learning provides the model with conditional probabilities similar to the ground truth answer for imitation without changing the probability model $P_{model}$. Specifically, an example problem \( q' \) and its corresponding correct solution \( r' \) is provided and it can be posited that the conditional probability \( P(r' \mid q') \) is similar to the probability of the ground truth answer of the target problem \( P(r_{gt} \mid q) \). Consequently, the model will imitate this similar example and \( r'_{predict} = \arg\max\limits_r P_{model}(r \mid q, q', r') \) will be closer to  \(r_{gt}\) comparing to \(r_{predict}\).

However, given that the actual reasoning process \( r \) can be highly complex, the complete reasoning process is often divided into multiple steps \( s_{1}, s_2, \ldots \). Step-level reasoning iteratively guides the model to generate the next step \( s^{0-shot}_{i+1} = \arg\max\limits_s P_{model}(s \mid q, s_1, s_2, \ldots, s_i) \). 

At the step granularity, examples retrieved based on the problem \( q \) are evidently insufficient for providing appropriate guidance. Similar problem \( q' \) may not necessarily contain the corresponding steps to guide the reasoning for the new problem \( q \). Moreover, irrelevant steps may provide dissimilar conditional probabilities, thereby distracting the model's reasoning process.

To this end, we propose step-aligned in-context learning and a first-try strategy to provide detailed and relevant example steps when in step-level reasoning. Specifically, when generating new steps \(s_{i+1}\) based on previous reasoning steps \(s_i, s_{i-1},\ldots, s_1\) and question \(q\), we first utilize a first-try strategy to obtain an approximate estimate of \( s^{first}_{i+1}\). Then, we use this \(s^{first}_{i+1}\) to retrieve a similar step \( s'_{n+1} \) along with the corresponding \( q', s'_1, s'_2, \ldots, s'_n \). Since these two steps are similar, a very reasonable assumption is that \(P(s'_{n+1} \mid q', s'_1, \ldots, s'_n) \) closely approximates \(P(s_{{gt}_{i+1}} \mid q, s_1, \ldots, s_i) \). Therefore, the generated step \(s_{i+1} = \arg\max\limits_sP_{model}(s \mid q, s_1, \ldots, s_i, q', s'_{1}, \ldots, s'_n, , s'_{n+1}, )\) will be more closed to \(s_{{gt}_{i+1}}\) comparing to \(s^{0-shot}_{i+1}\). Details about our step-level in-context learning and first-try strategy will be explained in Sec.~\ref{sec:retrieve}

\subsection{Step-Level Example Problem Bank}
\label{sec:data}
\begin{figure}[t]
\centering
\includegraphics[width=0.48\textwidth]{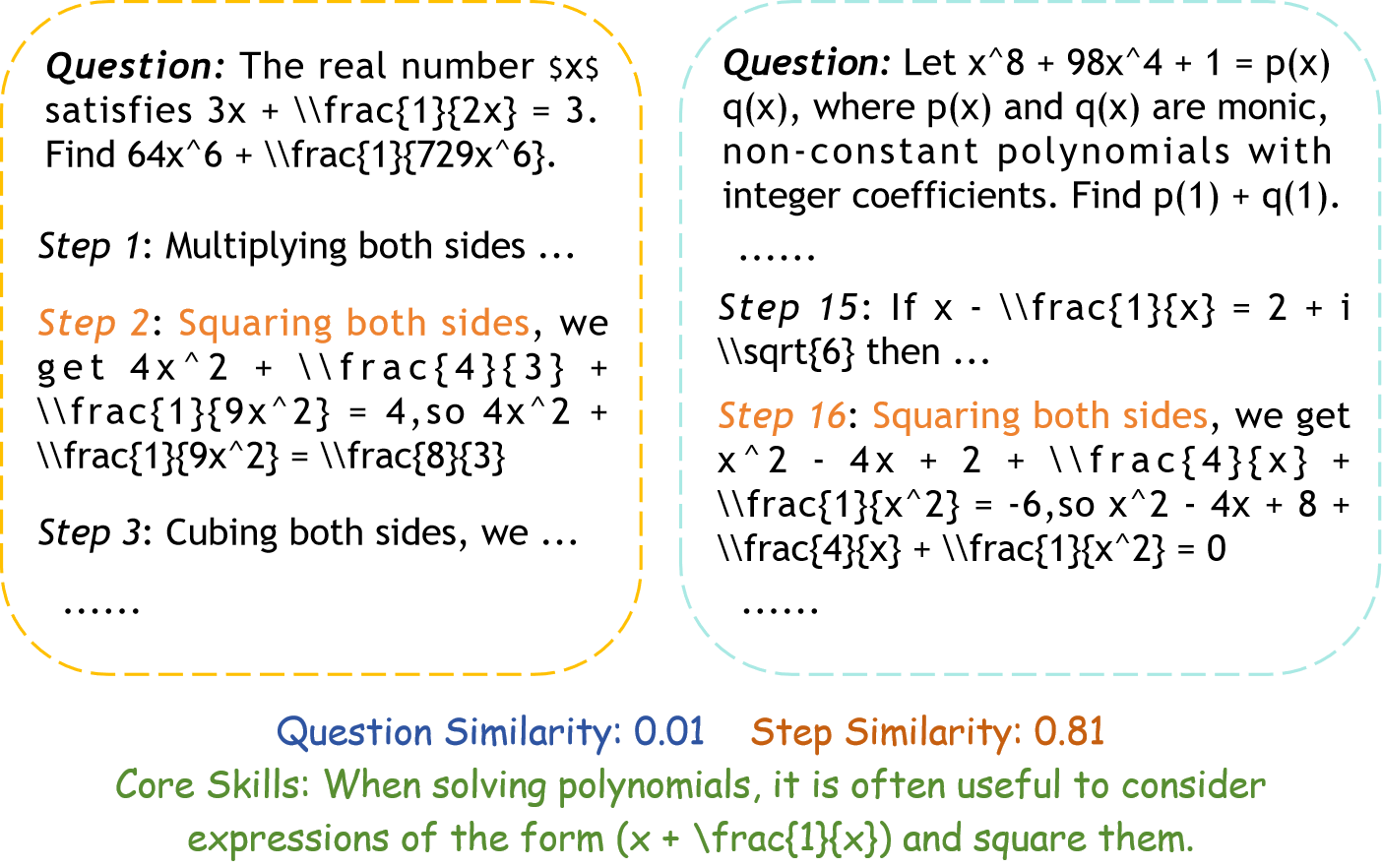}
\caption{Different problems may contain similar steps. Problem-level in-context learning will ignore this example due to low problem similarity. In contrast, our step-level in-context learning strategy can introduce the core skills by step-level retrieval and guidance.}

\label{fig:example}
\end{figure}

Due to the need for further improvement in mathematical capabilities, current open-source mathematical data no longer consist solely of problems and their final answers to determine whether the final answer obtained is correct or not. Instead, they also provide detailed solution processes to provide more fine-grained measurements. However, most current open-source mathematical data still do not break down the solution processes to the step level.

Some approaches~\cite{lightman2023let} proposed using a clear semantic delimiter like the period '.' or a new line to segment steps. This strategy allows for the quick decomposition of each step from a complete process without any additional assistance. However, this simple decomposition mode is obviously unreliable. Essentially, a single reasoning step should have a consistent target and a complete thought process, making it the atomic granularity of reasoning. Using a period '.' or a new line as a delimiter may disrupt this atomicity. For example, it may split a complete enumeration for the same objective into multiple steps.

Therefore, we suggest that the most appropriate method for step segmentation is to allow the reason model itself to autonomously decompose the process. This approach ensures that the granularity of the decomposed steps in example problem bank aligns with that of the real-time reasoning steps. Specifically, we define the concept of a step through prompts, which encapsulate a complete and simple inference. This guides GPT-4o in decomposing the answer at the step level.

A major advantage of decomposing the question example bank into individual steps is that it facilitates step-level retrieval and guidance, which is of significant importance. As illustrated in Fig.~\ref{fig:example}, two distinctly different problems may contain similar key steps. Traditional problem-level in-context learning often overlooks such examples, whereas step-level in-context learning can effectively recall these steps, thereby providing fine-grained guidance to the ongoing reasoning process$^{*}$.

\footnote{* The proposed step-level example problem bank is available at \url{https://github.com/beichenzbc/BoostStep}}

\subsection{Step-Level ICL with First-try Strategy}
\label{sec:retrieve}
The core challenge of in-context learning lies in how to effectively retrieve relevant problems or steps for effective guidance. This is contingent upon both the similarity between the problem database and the target problem, as well as the specific retrieval strategy employed. Traditional problem-level in-context learning involves retrieving similar problems based solely on the problem statement. This approach is relatively straightforward but effective, as similar problems typically encompass similar reasoning processes.

At the more granular step level, however, the situation becomes much more complex. A simple strategy is to perform retrieval using the given problem and all preceding reasoning steps \( s_{i-1}, s_{i-2}, \ldots, s_1, q \). The clear drawback of this method is the excessive length of the retrieval content, which diminishes the emphasis on the uniqueness of the current step. Another strategy is to use the previous step \( s_{i-1} \) to retrieve \( s'_{j-1} \) from a step-level database, thereby guiding the reasoning of \( s_i \) through the correct resolution of \( s'_j \). However, this approach is rather crude, as it models step-level reasoning as a Markov process, which is evidently unreasonable. Similar steps can be applicable to different reasoning tasks, and therefore similarity in the previous step does not necessarily indicate that the retrieved subsequent step will provide valuable guidance for the reasoning in the current step.

To this end, we propose a straightforward and effective "first-try" strategy to enhance the similarity of search steps. Our premise is that the most accurate way to estimate the next step is to actually allow the model to attempt the reasoning for the next step. Specifically, given a problem \( q \) and all preceding reasoning steps \( s_{i-1}, s_{i-2}, \ldots, s_1 \), we first instruct the model to attempt continuing the reasoning process to arrive at a tentative step \( s_i^{try} \) without the aid of any examples. Subsequently, we use \( s_i^{try} \) to retrieve similar steps \( s'_j \) along with their corresponding problem \( q' \) and preceding steps \( s'_1, \ldots, s'_{j-1} \) from a step-level database. Finally, we feed the retrieved similar steps back to the model, enabling it to deduce the final step \( s_i \). Besides, we add a widely accepted strategy reference rejection. Specifically, if the similarity of the retrieved most similar example remains below a certain threshold, we consider that there are no sufficiently similar examples available for reference. Consequently, we do not provide any examples to avoid the negative effects associated with incoherent in-context learning. This "try-retrieve-reason" strategy significantly enhances retrieval relevance, thereby improving reasoning effectiveness. Experiments in Sec.~\ref{sec:ablation} compare our method with several other retrieval strategies, demonstrating the superiority of our approach.

\subsection{Step-Level Guidance in Tree Search}
Our step-level in-context learning can significantly enhance the model's single-step reasoning capability, which makes it easily integrated into common step-level tree-search strategies. 

Generally, tree search methods necessitate two key components: a reason model that generates step-level reasoning and a Process-Supervised Reward Model (PRM) that continuously evaluates the current reasoning step in real time. Our method is beneficial for both of these components. It enhances the step-level reasoning performed by the reason model and improves the effectiveness of the PRM in evaluating current reasoning steps.

For the reason model, tree search methods inherently require step-by-step reasoning expansion. When expanding at node \( s_i \), we can apply the previously mentioned strategy: the model performs \( n \) first tries and retrieve for \( n \) example steps. For each example, the model then completes the reasoning to generate \( n \) child nodes \( s^1_{i+1}, \ldots, s^n_{i+1} \) with the help of these examples. Similarly, our strategy can improve the accuracy of individual nodes \( s^j_{i+1} \).

Evidently, judgment ability is closely related to reasoning ability. Therefore, since our strategy can enhance the accuracy of single-step reasoning, a reasonable assumption is that introducing appropriate example steps can improve the PRM's ability to assess the correctness of the current reasoning process. In particular, when evaluating the correctness of an inference step candidate \( s^j_i \), we retrieve similar steps \( s'_k \) along with their corresponding preceding steps \( s'_{k-1}, \ldots, s'_1 \) and question \( q' \) from the step-level example bank. Similarly, the probability distributions \( P(s'_k | s'_{k-1}, \ldots, s'_1, q') \) and \( P(s_{gt_i} | s_{i-1}, \ldots, s_1, q) \) exhibit similarities. This resemblance aids in assessing the discrepancy between \( s^j_i \) and \( s_{gt_i} \), thereby enhancing the accuracy of the critic model's evaluations.

Detailed ablation experiments in Sec.~\ref{sec:exp-mcts} demonstrate that both strategies contribute positively to step-level tree search methods.

\label{sec:tree}

\section{Experiments}
\subsection{Experiment setting}
\label{sec:setting}
\textbf{Reasoning Model.} \ Our primary reasoning model is GPT-4o~\cite{hurst2024gpt}. To demonstrate the generality, we also conducted tests on Qwen2.5-Math-72B-Instruct~\cite{yang2024qwen2}. Moreover, current SOTA reasoning models Qwen-QwQ-32B~\cite{qwq-32b-preview} and DeepSeek-R1-671B~\cite{guo2025deepseek} were also included in our experiment.

\noindent \textbf{Evaluation Benchmark.} \ We tested our approach on several challenging open-source mathematical benchmarks, including MATH500~\cite{hendrycks2021measuring}, AQuA~\cite{ling2017program}, OlympiadBench-TO~\cite{he2024olympiadbench} and MATHBench~\cite{liu2024mathbench} College-level and High-level tasks. In addition, we manually collected a selection of problems from the AMC-10 and AMC-12 competitions to serve as even more challenging benchmarks$^{*}$. \footnote{ * The AMC test set is available at \url{https://github.com/beichenzbc/BoostStep}}
To simulate benchmarks with lower similarity to the example problem bank, we also conducted tests on MathVision~\cite{wang2024measuring} and MathVerse~\cite{zhang2025mathverse}, highly challenging multi-modal math benchmarks

\noindent \textbf{Example Problem Bank.} The example problem bank is obtained from PRM800K~\cite{lightman2023let} and the steps are divided by GPT-4o.

\noindent \textbf{Retriever.} \ We utilized the classic TF-IDF encoding method combined with cosine similarity as the retriever for all methods. The TF-IDF weight matrix is derived from the example problem bank because the impact of the newly generated step is negligible, and real-time calculation of TF-IDF would require a significant amount of time.

\noindent \textbf{Hyper-Parameters.} The temperature value is 0 in all the experiments except for step-level tree search, which needs some random sampling to generate different reasoning candidates, and the temperature value for tree search methods is set at 0.3. The reference rejection threshold is 0.7.

\noindent \textbf{Prompt.} Apart from some necessary guidance like step-level reasoning, we ensured that the prompts for each method were as similar as possible to make the comparison fairer. The specific prompts are listed in the supplementary materials.

\subsection{Comparing to Problem-Level ICL}
\label{sec:comparing}

\begin{table*}[t]
  \centering
  \caption{A comparison of different in-context learning strategies on different benchmarks on GPT-4o and Qwen2.5-Math-72B-Instruct. The example problem bank is constructed from PRM800K, so MATH500 is an in-domain benchmark while others are all out-domain benchmarks. The best results are in \textbf{bold}.}
  \resizebox{\textwidth}{!}{
    \begin{tabular}{c|c|c|c|c|c|c|c|c|c}
    \toprule
    \multirow{2}{*}{Model}&\multirow{2}{*}{Method} & \textit{in-domain}&\multicolumn{6}{c}{\textit{out-domain}}& \multirow{2}{*}{\textbf{\large{Avg}}} \\
     \cmidrule(lr){3-3} \cmidrule(lr){4-9}
     &  & MATH & AMC12 & AMC10 & AQUA & MathBench(C) & MathBench(H) & OlympiadBench\\
    \midrule
    \multirow{3}{*}{GPT-4o}&0-shot & 73.4 & 53.6 & 55.8 & 81.1 & 80.0 & 77.3 & 40.6&66.0\\
    &few-shot & 73.8 & 56.5 & 56.7 & 83.9 & 80.7 & 79.3 & 39.3&67.2 (+1.2)\\
    &\textbf{Ours} & \textbf{76.4} & \textbf{63.0} & \textbf{60.4} & \textbf{85.4} & \textbf{82.0} & \textbf{84.0} & \textbf{43.3} & \textbf{70.6 (+4.6)}\\
    \midrule
    \multirow{3}{*}{Qwen}&0-shot & 83.0 & 67.4 & 67.7 & 84.6 & 80.6 & 82.0 & 49.7 & 73.6\\
    &few-shot & 83.8 & 67.4& 66.8 & 85.0 & 81.3 & 82.7 & 49.9 & 73.8 (+0.2)\\
    &\textbf{Ours} & \textbf{85.2} & \textbf{69.2} & \textbf{69.6} & \textbf{86.6} & \textbf{82.7} & \textbf{84.7} & \textbf{52.7} & \textbf{75.8 (+2.2)}\\
    \bottomrule
    \end{tabular}
    \label{tab:overall}
  }
\end{table*}

We conduct a rigorous comparison of our step-level in-context learning and traditional problem-level few-shot learning in various aspects. For traditional problem-level few-shot learning, we set the shot number to 4, which is a common setting. 

\textbf{Performance } We compare the performance between traditional few-shot learning and our step-level in-context learning across multiple benchmarks and base models. The results are presented in Tab.~\ref{tab:overall}. Our step-level in-context learning achieves a general and significant improvement across various benchmarks compared to problem-level few-shot learning.

\begin{table}[t]
  \centering
  \caption{Comparison of different strategies in multi-modal mathematical benchmarks with lower similarity with our problem bank. Base models are all GPT-4o.}
  \resizebox{0.48\textwidth}{!}{
    \begin{tabular}{c|c|c|c}
    \toprule
    Method & MathVision & MathVerse & \textbf{Avg} \\
    \midrule
    0-shot & 30.6 & 53.2 & 41.9\\
    few-shot & 28.7 (-1.9) & 53.2 (0.0) & 41.0 (-0.9)\\
    \textbf{Ours} & \textbf{35.2 (+4.6)} & \textbf{54.2 (+1.0)} &\textbf{44.7 (+2.8)}\\
    
    \bottomrule
    \end{tabular}
    \label{tab:multimodal}
  }
\end{table}

\begin{table}[t]
  \centering
  \caption{Experiments on the sensitivity of the similarity between the question and the example problem bank. R\_t indicates that the examples are the t\_th similar without any rejection strategy.}
  \resizebox{0.48\textwidth}{!}{
    \begin{tabular}{c|c|c|c}
    \toprule
    Method & Math-level5 & AMC12 & AMC10\\
    \midrule
    0-shot & 50.7 & 53.6 & 55.8 \\
    \midrule
    few-shot R\_1 & 52.2 (+1.5)&	56.5 (+2.9) &56.7 (+0.9)\\
    few-shot R\_4 & 46.3 (-4.4) &	52.2 (-1.4)&	53.7 (-2.1)\\
    \midrule
    Ours R\_1 & 56.0 (+5.3) & 62.3 (+8.7) & 60.4 (+4.6)\\
    Ours R\_4 & 52.2 (+1.5) &	61.6 (+8.0)	& 58.1 (+2.3)\\
    
    \bottomrule
    \end{tabular}
    \label{tab:curve}
  }
\end{table}

\textbf{Potential } A key focus of in-context learning is determining how difficult a particular example can effectively guide the new problems, indicating the potential of these methods. Problem-level in-context learning faces significant challenges in leveraging simpler examples to enhance the model's reasoning performance on more difficult questions. However, our strategy offers guidance at the step level, thereby overcoming this upper limit. To validate this, we select SOTA reasoning models QwQ-32B-Preview~\cite{qwq-32b-preview} and DeepSeek R1~\cite{guo2025deepseek} and utilized simpler example problems from PRM800K to guide the reasoning on the most challenging mathematical benchmark AIME~\cite{AIME}. The results are shown in tab.~\ref{tab:aime}, which indicate that traditional few-shot learning fails to provide effective guidance while our strategy demonstrates continuous improvement, demonstrating that it can boost the most advanced reasoning models on the most challenging tasks with a much simpler example.

\begin{table}[t]
  \centering
  \caption{Experiment on "simple-aids-difficult" potential. We use simpler example problems from PRM800K to guide SOTA reasoning models Deepseek-R1 and Qwen-QwQ on the most challenging mathematical benchmarks AMC 12 and AIME. Considering that the AIME consists of only 30 questions each year, making the results prone to fluctuations, we evaluated the questions three times annually and reported the average accuracy.}
  \resizebox{0.48\textwidth}{!}{
    \begin{tabular}{c|c|c|c|c}
    \toprule
    Model & Method & AMC12 & AIME23 & AIME24 \\
    \midrule
    \multirow{3}{*}{QwQ} & 0-shot & 79.7 & 38.9 & 43.3 \\
    & few-shot & 81.2 & 33.3 (-5.6) & 38.9 (-4.4)\\
    & Ours & \textbf{88.4} & \textbf{41.1 (+2.2)} & \textbf{47.8 (+4.5)}\\
    \midrule
    \multirow{3}{*}{DS-R1} & 0-shot & 94.2 & 75.6 & 80.0 \\
    & few-shot & \textbf{97.1} & 65.6 (-10.0) & 70.0 (-10.0)\\
    & Ours & \textbf{97.1} & \textbf{77.8 (+2.2)} & \textbf{82.2 (+2.2)}\\
    \bottomrule
    \end{tabular}
    \label{tab:aime}
  }
\end{table}

\textbf{Generalizability } 
Traditional few-shot learning requires the example problem bank highly similar to the questions to be solved, which limits its generalizability. To compare the generalizability, We also test different methods on multi-modal mathematical benchmarks including MathVision~\cite{wang2024measuring} and MathVerse~\cite{zhang2025mathverse}, which has much lower similarity with our example problem bank. The results are shown in Tab.~\ref{tab:multimodal}. Problem-level few-shot learning not only fails to enhance reasoning performance but can also have a negative impact, while our method continues to achieve appreciable improvements, demonstrating better general applicability.

\textbf{Robustness } We also manually decrease the similarity between the examples and the problems by selecting the t\_th similar example during reasoning to evaluate the robustness. The result is shown in Tab.~\ref{tab:curve}. We can observe that traditional problem-level in-context learning suffers from a severe decrease and is even worse than 0-shot learning when t is larger than 4. In contrast, our method does not show a significant decline and is consistently better than the 0-shot reasoning.

\subsection{Construction of Example Problem Bank}
To better align with the steps in reasoning, we propose constructing a step-level problem bank based on the reasoning content rather than grammatical divisions. To prove our assumption, we compare our approach with a commonly used strategy that constructs steps based on grammatical segmentation, using periods \textbf{'.'} as the delimiter, on the same dataset PRM800K and under identical conditions. Results are presented in Tab.~\ref{tab:problem}.  Our method largely outperforms those using periods as a delimiter.

\begin{table}[t]
  \centering
  \caption{Comparison of different step-level example problem Bank construction methods.}
  \resizebox{0.48\textwidth}{!}{
    \begin{tabular}{c|c|c|c}
    \toprule
    Strategy & AMC12 & AMC10 & MATH \\
    \midrule
    Grammatical Separation & 56.5 & 58.1 & 74.8 \\
    \textbf{Reasoning Content} & \textbf{63.0} & \textbf{60.4} & \textbf{76.4}\\
    
    \bottomrule
    \end{tabular}
    \label{tab:problem}
  }
\end{table}

\subsection{Comparison of Retrieving Strategies}
\label{sec:ablation}
The key factor of in-context learning lies in the relevance of the retrieved examples. At the finer-grained step level, designing an appropriate retrieval strategy becomes even more crucial and challenging. Therefore, we propose the first-try strategy, which involves understanding what the model currently needs to reason about using a first attempt and then searching the problem set for similar steps to guide the model in fully outputting the current step. To validate the effectiveness of this method, we compare it with several other strategies mentioned in Sec.\ref{sec:retrieve}, retrieving by the entire reasoning path  \( s_{i-1}, s_{i-2}, \ldots, s_1 , q\) or only by the immediately preceding step \( s_{i-1}\).

Tab.~\ref{tab:retrieve} presents the detailed result. Our method significantly outperforms the other two retrieving strategies, better anticipating the content that needs to be inferred in the current step.

\begin{table}[t]
  \centering
  \caption{Comparison of different retrieval strategies in step-level in-context learning. 'Path' represents retrieving by the reasoning path including all previous steps \( s_{i-1}, s_{i-2}, \ldots, s_1 \) and question \(q\), while 'Pre-Step' represents retrieving by only the immediately preceding step \( s_{i-1}\). The best results are in \textbf{bold}.}
  \resizebox{0.48\textwidth}{!}{
    \begin{tabular}{c|c|c|c|c}
    \toprule
     Strategy & AMC12 & AMC10 & MATH & MathVision\\
    \midrule
    Path & 56.5 & 58.1 & 73.8 & 31.7\\
    Pre-Step & 57.2 & 56.7 & 74.0 & 31.0\\
    \textbf{First-try} & \textbf{63.0} & \textbf{60.4} & \textbf{76.4} & \textbf{35.2}\\
    
    \bottomrule
    \end{tabular}
    \label{tab:retrieve}
  }
\end{table}

\subsection{Example-guided Step-level Tree Search}
\label{sec:exp-mcts}

\begin{table}[t]
  \centering
  \caption{A detailed ablation on incorporating retrieving similar steps to provide fine-grained guidance during the reasoning and verifying phases of step-level tree search methods. Base models are GPT-4o and prompts are the same. The best results are in \textbf{bold.}}
  \resizebox{0.48\textwidth}{!}{
    \begin{tabular}{c|c|c|c|c|c}
    \toprule
     Reason & Verify & AMC12 & AMC10 & MATH & Avg\\
    \midrule
    \multicolumn{2}{c|}{w/o tree-search} & 53.6 & 55.8 & 73.4 & 60.9 \\
    \midrule
    \XSolidBrush & \XSolidBrush & 58.7 & 59.0 & 77.8 & 65.2 (+4.3)\\
    \Checkmark & \XSolidBrush & 64.4 & 62.2 & 79.2 & 68.6 (+7.7)\\
    \XSolidBrush & \Checkmark & 61.6 & 60.4 & 78.2 & 66.7 (+5.8)\\
    \Checkmark & \Checkmark & \textbf{65.2} & \textbf{63.6} & \textbf{79.4} & \textbf{69.4 (+8.5)}\\
    
    \bottomrule
    \end{tabular}
    \label{tab:mcts}
  }
\end{table}

The reasoning capability of the reason model and the verifying capability of the critic model are two core factors of step-level tree search methods, and our strategy can bring benefits in both ways. On one hand, it can improve the accuracy of generating candidate nodes using the previously mentioned first-try strategy when reasoning nodes are generated. On the other hand, it can increase the accuracy of evaluation by introducing similar examples during critic model assessments and therefore ensures that the correct reasoning nodes are more likely to be preserved. These can be decoupled, allowing us to demonstrate the effectiveness of each component through ablation studies.

We utilize GPT-4o as the reason model, GPT-4o-mini as the PRM and adopt the Pairwise Preference Reward Model (PPRM) configuration~\cite{zhang2024llama} to ensure a more robust evaluation. Detailed settings will be listed in the appendix. 

Tab.~\ref{tab:mcts} presents the results of integrating in-context learning into the reasoning and evaluation phases of Tree Search methods. The results of this ablation study indicate that introducing example steps can enhance both the reasoning and verifying capabilities of tree search methods. Therefore both approaches contribute to the improvement of overall reasoning performance.

\section{Conclusion}
We propose BoostStep, providing fine-grained guidance during the reasoning process by searching for similar steps from a step-level example problem bank according to the first-try reasoning attempt. BosotStep is a strong and general approach, enhancing the model's reasoning capabilities and reducing the dependency on the similarity of the example problem set. It demonstrates better performance, potential, generalizability and robustness comparing to traditional problem-level few-shot learning. Moreover, our method can also enhance the reasoning and evaluation capability of step-level tree search methods by introducing similar steps in reasoning and verifying phases.

\section{Limitations}

Currently, our example problem bank is entirely sourced from PRM800k, resulting in a relatively homogeneous distribution of example problems and example steps. Although our method has more potential to guide more difficult problems with much simpler examples, a greater quantity and more diverse distribution of example problems can evidently provide more effective guidance for addressing a range of problems. 

Furthermore, the TF-IDF retriever used is based on modeling language term frequency directly and thus lacks an understanding of mathematical content, which limits its retrieval capabilities on math problems. Utilizing a retriever specifically designed for mathematical problems can certainly enhance the quality of retrieval.

\section*{Acknowledgments}

This project is funded in part by Shanghai Artificial lntelligence Laboratory, the National Key R\&D Program of China (2022ZD0160201), the Centre for Perceptual and Interactive Intelligence (CPII) Ltd under the Innovation and Technology Commission (ITC)’s InnoHK. Dahua Lin is a PI of CPII under the InnoHK.

\bibliography{custom}

\clearpage
\appendix

\section{Detailed Experiment Setting}
\label{sec:appendix}
\subsection{Prompt}
\textbf{Prompt for 0-shot COT}: You are a professional math problem solver. Solve the problem step by step and output the final answer within \textbackslash \textbackslash boxed\{\}.
\\

\noindent \textbf{Prompt for problem-level few-shot learning}: You are a professional math problem solver.  Solve the problem step by step and output the final answer within \textbackslash \textbackslash boxed\{\}. In case you don't know how to solve it, I will give you example problems with their full solutions which you can refer to.

\noindent Example i: 

\noindent Problem: xxx  

\noindent Solution: xxx  
\\

\noindent \textbf{Prompt for first-try in step-level COT}: You are a professional math problem solver. I will give you a math problem and part of its solution. And you need to only output the next step of the solution, starting with 'Step $i$:', where $i$ is the step number. If you think that the final step is derived, put the answer within \textbackslash \textbackslash boxed\{\}.
\\

\noindent \textbf{Prompt for step-level few-shot learning}: You are a professional math problem solver. I will give you a math problem and part of its solution. And you need to only output the next step of the solution, starting with 'Step $i$:', where $i$ is the step number. In case you don't know how to derive the correct content, an example with 'Key Step' will be given. You need to learn how 'Key Step' is derived, and implement similar strategy in your derivation procedure. If you think that the final step is derived, put the answer within \textbackslash \textbackslash boxed\{\}. 

\noindent Example Problem: xxx 

\noindent Example Solution: Step1: xxx, Step2: xxx, ..., Stepi(Key Step): xxx. 

\subsection{Details of Grading and Metrics}
We follow the setting of Opencompass~\cite{2023opencompass} and VLMEvalKit~\cite{duan2024vlmevalkit}. Specifically, we first require the model to put the final answer within \textbackslash \textbackslash boxed\{\}. Then, we use GPT-4o-mini as the critic model to compare the final answer with the ground truth answer. Compared to string matching, this approach can eliminate some false negative evaluations because the same mathematical expression can be expressed in many forms. If the model fails to follow the
the expected format in the prompt and the rule-based
extraction fails, the solution is directly judged as inconsistent with ground truth.

\subsection{Benchmarks}
We tested our approach on several mathematical benchmarks, including MATH500~\cite{hendrycks2021measuring}, AQuA~\cite{ling2017program}, OlympiadBench-TO~\cite{he2024olympiadbench} and MATHBench~\cite{liu2024mathbench}. Specifically, we use the Olympiad-TO (text-only) subset of OlympiadBench and the application problems in college-level and high-level difficulty of MATHBench.

For multi-modal math benchmarks, we use MathVision-Mini~\cite{wang2024measuring} and vision-dominant version of problems in MathVerse-Mini ~\cite{zhang2025mathverse}.

\section{Detailed Setup for Example-Guided Step-Level Tree Search}
In the setup for tree search methods, we utilize GPT-4o as the reason model and employ GPT-4o-mini as the Process-supervised Reward Model (PRM). For the PRM, we adopted the Pairwise Preference Reward Model (PPRM) configuration~\cite{zhang2024llama}. Specifically, PPRM transforms the absolute rewards calculation into preference predictions between solutions to calculate rewards. This approach reduces the variability associated with scoring characteristics and thus leads to a more robust and consistent evaluation of different solutions.

The complete reasoning process in our experiment is as follows: we start with the target problem as the root node and obtain two initial solution steps through sampling to serve as the two initial parent nodes. In each step-level reasoning phase, we expand these two parent nodes through sampling, generating four candidate child nodes. Using the PPRM, we select the two child nodes with higher confidence to become the parent nodes for the next step of reasoning. This process continues until both candidate nodes have completed their reasoning paths, resulting in the final answers. Finally, PPRM is used to select the ultimate answer from these two reasoning paths.

\section{Case Study}

\begin{figure}[t]
\centering
\includegraphics[width=0.48\textwidth]{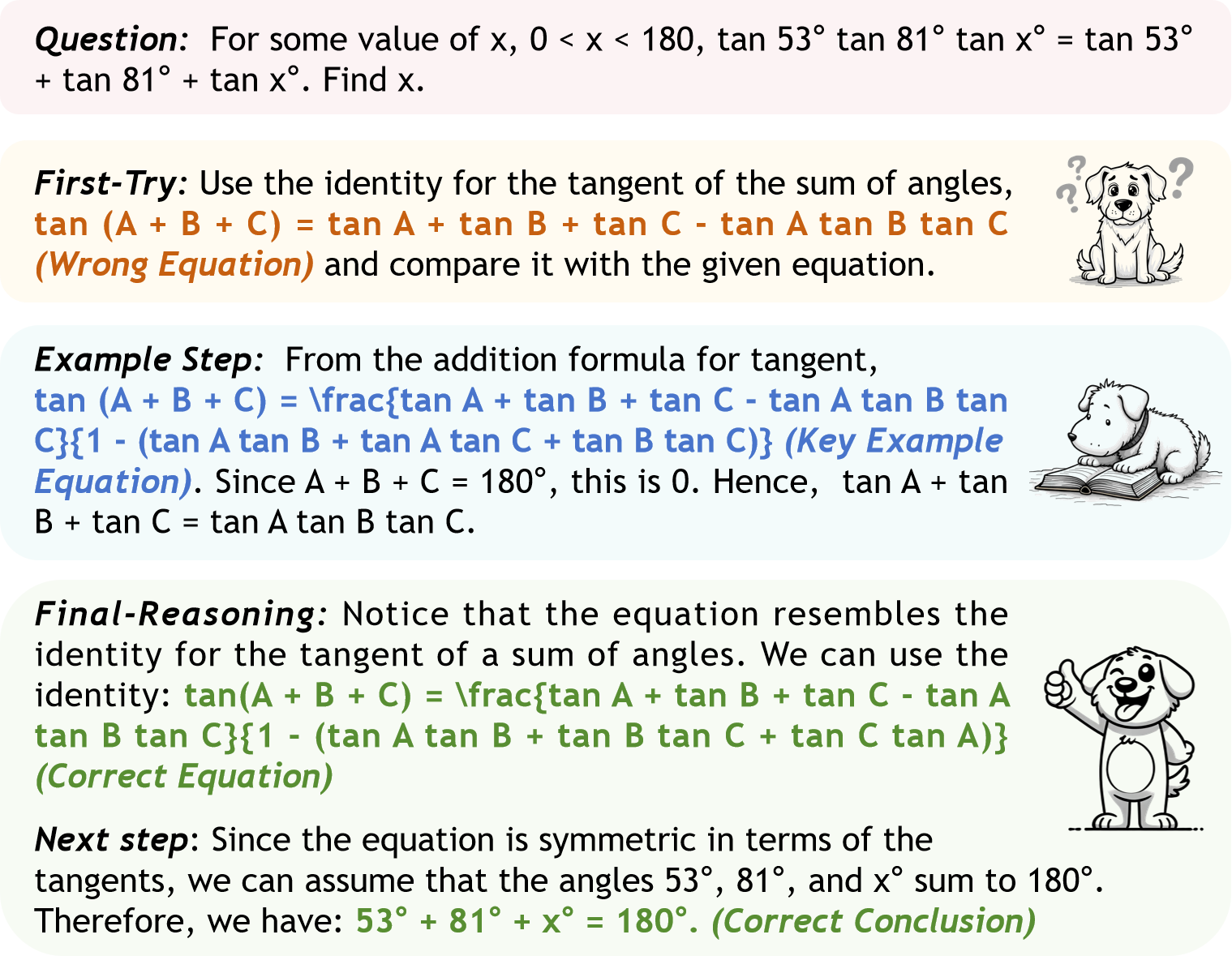}
\caption{A specific example of adjusting reasoning during real-time inference through step-level in-context learning. The first try uses a wrong equation while the retrieving example step guides the model to use the correct equation and get the correct conclusion.}

\label{fig:case}
\end{figure}

Here we demonstrate a specific example of how our step-level in-context learning boosts step-level reasoning. Given the question, we first let the model have a first try on step one. Unfortunately, because the model is unfamiliar with trigonometric functions, it makes an error on the tangent sum formula, therefore leading to a wrong step. However, we can get a rough idea of what the model wants to calculate at this step according to the first try. Then, we find a similar step that correctly leverages the tangent sum formula in the step-level example problem bank. Therefore, with the guidance provided, the model correctly applied the tangent sum formula during the second reasoning attempt and arrived at the correct answer. 

\end{document}